\documentclass[runningheads]{llncs}
\usepackage{graphicx}
\usepackage{caption}
\usepackage{cite}
\usepackage{subcaption}
\usepackage{tabularx}
\usepackage{xcolor}
\usepackage{placeins}

\begin{document}
%
\title{An NLP approach to quantify dynamic salience of predefined topics in a text corpus\thanks{This material is based upon work supported by the Small Business Innovative Research/Small Business Technology Transfer (SBIR/STTR) Program and the Engineering Research and Development Center (ERDC) – Construction Engineering Research Laboratory (CERL) under Contract W9132T20C0012. Any opinions, findings and conclusions or recommendations expressed in this material are those of the author(s) and do not necessarily reflect the views of the SBIR/STTR Program and ERDC-CERL.\newline\newline This paper was presented at the 2021 International Conference on Social Computing, Behavioral-Cultural Modeling \& Prediction and Behavior Representation in Modeling and Simulation (SBP-BRiMS), 9 July 2021.}}
\titlerunning{Dynamic salience of predefined topics}
%
\author{A.~Bock\inst{1}\orcidID{0000-0003-1870-8499} \and
A.~Palladino\inst{1}\orcidID{0000-0002-2697-3869} \and
S.~Smith-Heisters\inst{2}\orcidID{0000-0003-3447-0572} \and
I.~Boardman\inst{1}\orcidID{0000-0002-4588-9400} \and
E.~Pellegrini\inst{1}\orcidID{0000-0001-6203-8774} \and
E.J.~Bienenstock\inst{2}\orcidID{0000-0001-8747-8872} \and
A.~Valenti\inst{1}\orcidID{0000-0001-6611-552X}}
\authorrunning{A. Bock et al.}
%
\institute{Boston Fusion Corp., Lexington, MA 02421, USA \and
Arizona State University, Phoenix, AZ 85004, USA}
%
\maketitle              
\begin{abstract}
The proliferation of news media available online simultaneously presents a valuable resource and significant challenge to analysts aiming to profile and understand social and cultural trends in a geographic location of interest. While an abundance of news reports documenting significant events, trends, and responses provides a more democratized picture of the social characteristics of a location, making sense of an entire corpus to extract significant trends is a steep challenge for any one analyst or team. Here, we present an approach using natural language processing techniques that seeks to quantify how a set of predefined topics of interest change over time across a large corpus of text. We found that, given a predefined topic, we can identify and rank sets of terms, or $n$-grams, that map to those topics and have usage patterns that deviate from a normal baseline. Emergence, disappearance, or significant variations in $n$-gram usage present a ground-up picture of a topic’s dynamic salience within a corpus of interest.

\end{abstract}
\section{Introduction}

Contemporary researchers interested in measuring the social characteristics and dynamics of a given environment are presented with a massive amount of accessible digital information from which they can arrive at answers to their questions. The amount of written news material alone available online provides a valuable wealth of information, background, and documentation that can be used to answer social scientific questions. Sifting through this trove of data and highlighting discussions and trends of particular interest proves a challenge, even at the computational level, let alone for individual researchers to pursue.

Purely unsupervised identification of useful material from online news corpora is one approach. Nevertheless, semantic frameworks have been developed, particularly in the defense and intelligence domains (Table~\ref{tab1}), to identify particular topics of interest and sort findings accordingly. Nation-state assessment (DIME-FIL)~\cite{hartley-2015}, infrastructure assessment (SWEAT-MSO)~\cite{army-2008}, situation assessment (METT-TC)~\cite{ulicny-2007}, operational variables (PMESII-PT)~\cite{hartley-2015}, and civil considerations (ASCOPE)~\cite{army-2010} are examples of frameworks into which content can be organized for specific analyses. Superimposing such frameworks on findings extracted from corpus analysis frames the analysis within standardized contexts and facilitates knowledge sharing.

In this study, we aim to describe the results of term frequency analysis across a corpus in terms of the PMESII-ASCOPE framework, a hybrid organizational structure of the aforementioned PMESII and ASCOPE frameworks (see Table~1). Crucially, we aim to measure how the salience of PMESII-ASCOPE topics, derived from observed terms occurring in the corpus, may change over time and correlate over time with respect to known events discussed by the corpus’ constituent documents. Compared to a survey of existing related approaches, our analysis borrows from and builds upon existing methods to offer three primary strengths in an innovative approach to addressing the stated problem: Capturing meaningful language use above the level of the word (i.e., the $n$-gram, or a distinct ordering of $n$ tokens identified by a lexer), the use of predefined topics to empower framework-based analysis, and considering temporal change in $n$-gram usage across a corpus of documents over a time period of interest. Kherwa and Bansal~\cite{kherwa-2019}, though using a topic modeling approach, similarly harness the semantic power of $n$-grams; they cite the difference between the words ``New" and ``York" occurring separately and ``New York" as a bigram. (``Black Lives Matter" is another example with particular relevance to the PMESII-ASCOPE space.) We apply a term-frequency measurement similar to Don et al.~\cite{don-2007}, but scaled to the level of a corpus, as opposed to single documents. Ahmed, Traore, and Saad~\cite{ahmed-2017} note an epistemological limitation in their analysis: Identification of terms and trends of interest in text data typically requires some a priori knowledge of what is to be found. While topic modeling is useful for discovering latent topics that can be later associated with known trends or events, our approach empowers the reverse analysis; i.e., we hypothesize that, given a known set of topics, we can discover emergent trends or events by tracking dynamic topic salience. We synthesize insights from aforementioned research and develop a novel approach to test this hypothesis.

\begin{table}[h]
\caption{Common topical frameworks and their uses.}\label{tab1}
\begin{tabular}{l|l}
\hline
{\bf Topical Framework: Predefined Topics} &  {\bf Utility} \\
\hline
\begin{minipage}{3.0in}
\vspace{0.25cm}
DIME-FIL: Diplomatic, Information, Military, Economic, Financial, Intelligence, and Law Enforcement
\end{minipage} &  Nation-state power assessment\\
\begin{minipage}{3.0in}
\vspace{0.25cm}
SWEAT-MSO: Sewer, Water, Electricity, Academics, Trash, Medical, Safety, and Other
\end{minipage} &  Infrastructure assessment\\
\begin{minipage}{3.0in}
\vspace{0.25cm}
METT-TC: Mission, Enemy, Terrain, Troops, Time, and Civilians
\end{minipage} & Situation assessment\\
\begin{minipage}{3.0in}
\vspace{0.25cm}
PMESII-PT: Political, Military, Economic, Social, Infrastructure, Information, Physical environment, Time
\end{minipage} & Operational variables\\
\begin{minipage}{3.0in}
\vspace{0.25cm}
ASCOPE: Areas, Structures, Capabilities, Organizations, People, Events
\end{minipage} & Civil considerations\\
\hline
\end{tabular}
\label{tab1}
\end{table}

\FloatBarrier
\section{Defining $n$-gram usage}

To enable temporal term-based analysis of a corpus, the temporal usage of unique terms in the corpus must be quantified. We use $n$-grams (where $n$=2 for this paper) occurring across a corpus as base units of analysis. First, we construct the set of all $n$-grams that appear in the corpus by tokenizing each document and counting usages of unique $n$-grams. We then binned documents by month of publication; in practice, time bin size could be chosen based on the frequency of documents under investigation. For each $n$-gram, $g$, over $m$ time units, we generated a relative usage trend of $m$ values by calculating the proportion of observed $n$-grams in each time bin that are instances of $g$.

We used the Big Open-Source Social Science (BOSSS) tool~\cite{palladino-2018, perkins-2020} to collect a corpus of 2,760 news articles from Google News discussing Malindi, a city on the Kenyan coast, published between January 2016 and September 2018. As a validation technique for our usage quantification, we identified particular events the corpus discusses that we expected to associate with use of certain $n$-grams and ultimately with PMESII-ASCOPE topics. The corpus' time frame includes the August 2017 Kenyan general election, its judicial annulment, a subsequent election in October 2017, and an opposition party event in January 2018. We invoke these events as we consider usage trends for select $n$-grams and later salience trends for select PMESII-ASCOPE topics.

\begin{figure}[h]
\centering
     \begin{subfigure}[b]{0.49\textwidth}
         \centering
         \includegraphics[width=\textwidth]{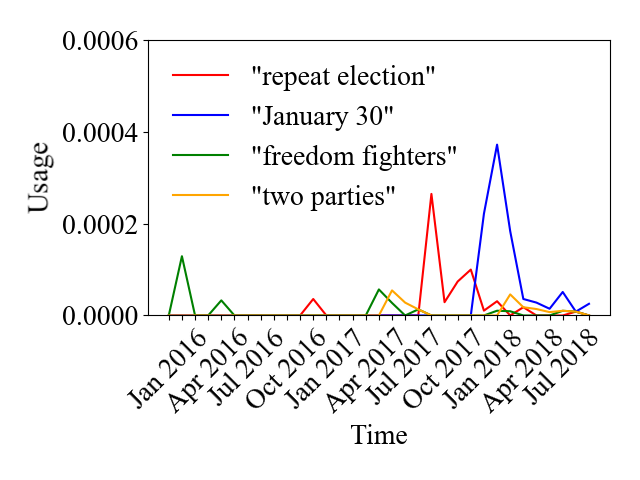}
         \caption{`Political Events'}
         \label{fig:fig1a}
     \end{subfigure}
     \hfill
     \begin{subfigure}[b]{0.49\textwidth}
         \centering
         \includegraphics[width=\textwidth]{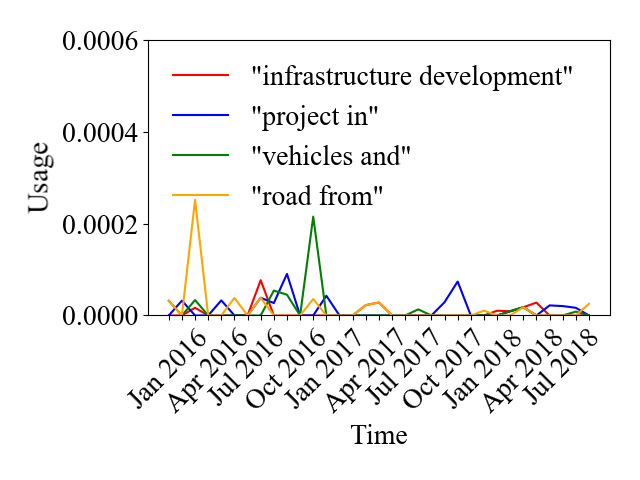}
         \caption{`Infrastructure Capabilities'}
         \label{fig:fig1b}
     \end{subfigure}
\vspace{-0.25cm}
\caption{Relative usage trends of $n$-grams associated with two topics. Note that elections occurred in both August and October 2017.
}
\label{fig1}
\end{figure}

Figure 1 presents relative usage trends for selected $n$-grams associated with the ``political events" and ``infrastructure capabilities" PMESII-ASCOPE topics. Expectedly, we observed an uptick in use of electoral $n$-grams (“polls were”, “Jubilee supporters”) preceding and during the time bins in which the elections took place. Importantly, we note that our usage quantification method can identify the emergence of novel $n$-grams in the corpus. Prior to August 2017, we observe comparatively little discussion of a “repeat election”; after the first round was held, usage of this $n$-gram spiked, rising again in the time bins following the October 2017 repeat election and continuing but declining in usage after the event. This suggests little precedent for journalistic interest in a ``repeat election" until the wake of the first round, when an imminent repeat election became apparent. Similarly, our algorithm detects no discussion of ``January 30" until after the October 2017 repeat election; the opposition party hosted a symbolic inauguration of its candidate on 30 January 2018, in response to the results of the two elections held. In Figure~\ref{fig:fig1b} we show the usage of example $n$-grams mentioned in context with the ``infrastructure capabilities" topic; note that while discussion of all $n$-grams tracked is observed throughout the time frame, a cluster of mutual usage is observed surrounding August 2016. These trends suggest notable correlation between events occurring during the corpus’ time frame and dynamic use of $n$-grams with presumed relationships to those events. In addition, they demonstrate that this method is responsive to $n$-grams that appear suddenly and with significant usage (e.g. ``repeat election", ``January 30") and is able to discover terms that may arise without precedent, particularly important for understanding the impacts of sudden or unexpected events without a priori knowledge on social discussion and journalistic output.

\FloatBarrier
\section{$N$-gram–topic similarity scoring and mapping}

Unlike topic modeling techniques commonly used in corpus analysis, e.g., Latent Dirichlet Allocation (LDA)~\cite{blei-2003, chuang-2012}, this study uses predefined topics with ground-truths defined a priori rather than discovering topics. We built a vector-space model for each predefined topic based on a ground-truth collection of text, relevant keywords, and topic definitions. We expanded the topic definition to include all known synonyms of included terms in the definition, then applied term frequency - inverse document frequency (TF-IDF) to assist in distinguishing between topics, resulting in a unique vector-space model for each of the PMESII-ASCOPE topics described in Section~1.

We aimed to associate to each topic a set of observed $n$-grams whose usage related to a discussion of that topic within the corpus. For each $n$-gram, we extracted the $n$-gram’s contexts, or the phrases in which each instance of the $n$-gram occurs. Using the same vector space as our topic models, we generated vector representations of each $n$-gram (by averaging the vector representations of each of their contexts). Using these representations, we calculated the cosine similarity between vector representations of each $n$-gram and each PMESII-ASCOPE topic, represented as a similarity matrix.
\begin{figure}[htb]
     \centering
     \begin{subfigure}[b]{0.49\textwidth}
         \centering
         \includegraphics[width=\textwidth]{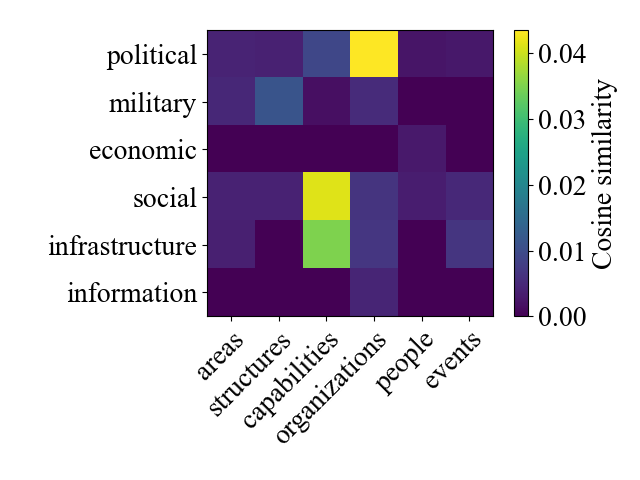}
         \vspace{-0.9cm}
         \caption{`education system'}
         \label{fig:fig2a}
     \end{subfigure}
     \hfill
     \begin{subfigure}[b]{0.49\textwidth}
         \centering
         \includegraphics[width=\textwidth]{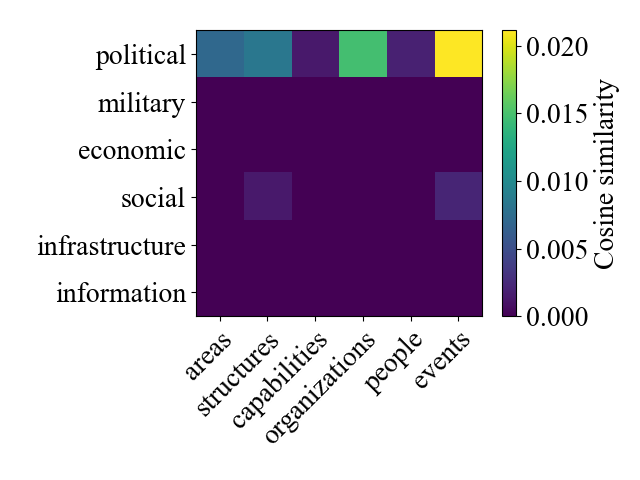}
         \vspace{-0.9cm}
         \caption{`polls were'}
         \label{fig:fig2b}
     \end{subfigure}
     \hfill
     \begin{subfigure}[b]{0.49\textwidth}
         \centering
         \includegraphics[width=\textwidth]{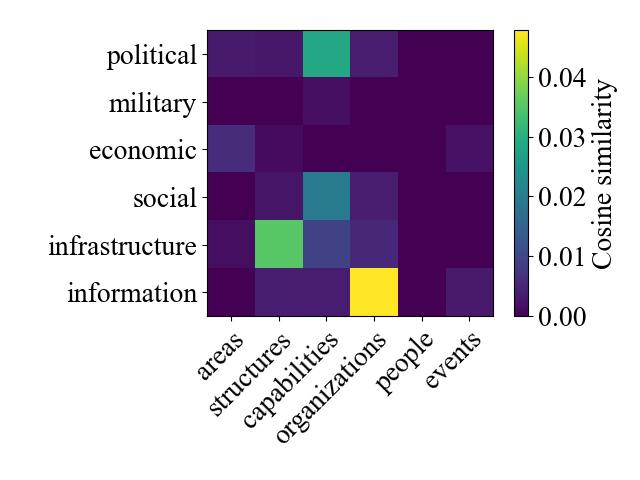}
         \vspace{-0.9cm}
         \caption{`public hospitals'}
         \label{fig:fig2c}
     \end{subfigure}
     \begin{subfigure}[b]{0.49\textwidth}
         \centering
         \includegraphics[width=\textwidth]{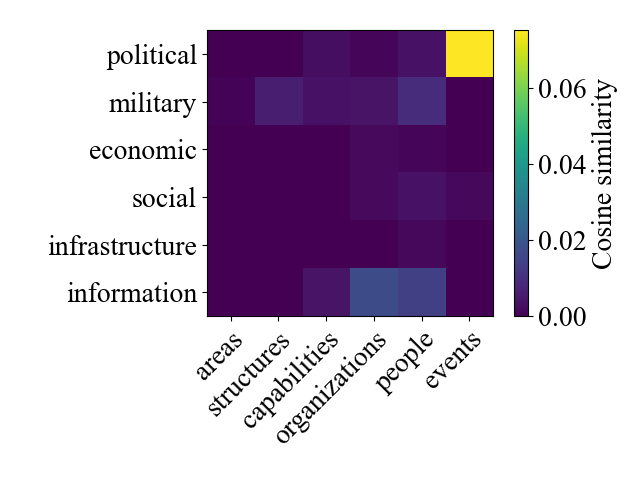}
         \vspace{-0.9cm}
         \caption{`repeat election'}
         \label{fig:fig2d}
     \end{subfigure}
     \vspace{-0.25cm}
        \caption{Similarity matrices of example $n$-grams quantifying the topical relevance of the context in which the $n$-grams are used. PMESII categories are the rows, ASCOPE categories are the columns, and each cell is a topic.}
        \label{fig:fig2}
\end{figure}

\noindent Figure 2 demonstrates the results of topic similarity scoring for four example bigrams. The matrix generated for the $n$-gram “repeat election” is heavily skewed to just the one “political events” \underline{P}MESII-ASCOP\underline{E} topic, while the similarity calculated for “polls were” maps significantly to categories across the political row. The $n$-grams “education system” and “public hospitals” also show notable similarity to infrastructural topics, as expected, but additionally for political, social, and information categories. Less skewed, more evenly distributed similarity matrices may suggest discussion of an $n$-gram in multiple semantic contexts, providing additional and perhaps unexpected context to downstream salience quantification and to individual $n$-gram–topic relationships.

For each $n$-gram, we quantified variability in its usage by calculating the relative standard deviation of the values constituting its relative usage trend. For each topic, the $n$-grams that exceed the 75$^{\mathrm{th}}$ percentile of both variability in usage and similarity to the topic are associated with that topic, meaning we use them to analyze that topic’s salience (as described in Section~5). Using this bivariate analysis, we implemented a discriminatory $n$-gram–topic association method that both depends on strong topic similarity and captures maximal temporal fluctuation in usage of topic-associated $n$-grams. Figure~3 illustrates the bivariate topic association measure for the “political events” topic (whose most similar associated $n$-grams are listed in Table 2). Note, we do not filter out stop words or mundane phrases, such as ``have reached" (9$^\mathrm{th}$ place in Table~2), to enable our approach to capture emergent socially-relevant phrases such as ``me too", i.e., \#metoo.
\vspace{-0.5cm}
\begin{table}[h!]
\caption{Top 10 most similar $n$-grams associated with the ``political events" topic. Similarity values are one element of the bivariate association.}
\label{tab2}
\centering
\begin{tabularx}{\textwidth}{p{0.5\textwidth} | p{0.5\textwidth}}
\hline
\bf{$n$-gram} & \bf{Similarity} \\
\hline
`repeat election' & 0.0753 \\
`election which' & 0.0750 \\
`our security' & 0.0401 \\
`event to' & 0.0294 \\
`October 2017' & 0.0259 \\
`challenging the' & 0.0230 \\
`polls were' & 0.0212 \\
`social and' & 0.0157 \\
`have reached' & 0.0121 \\
`s victory' & 0.0120 \\
\hline
\end{tabularx}
\end{table}
\vspace{-1.5cm}
\begin{table}[h!]
\caption{Top 10 most similar $n$-grams associated with the ``infrastructure areas" topic. Similarity values are one element of the bivariate association.}
\label{tab3}
\centering
\begin{tabularx}{\textwidth}{p{0.5\textwidth} | p{0.5\textwidth}}
\hline
\bf{$n$-gram} & \bf{Similarity} \\
\hline
`road from' & 0.0407 \\
`infrastructure development' & 0.0369 \\
`County the' & 0.0314 \\
`project will' & 0.0113 \\
`sit on' & 0.0104 \\
`total number' & 0.0093 \\
`residents who' & 0.0091 \\
`project in' & 0.0083 \\
`the lake' & 0.0077 \\
`is taking' & 0.0065 \\
\hline
\end{tabularx}
\end{table}

\begin{figure}[h]
\centering
\includegraphics[width=0.85\textwidth]{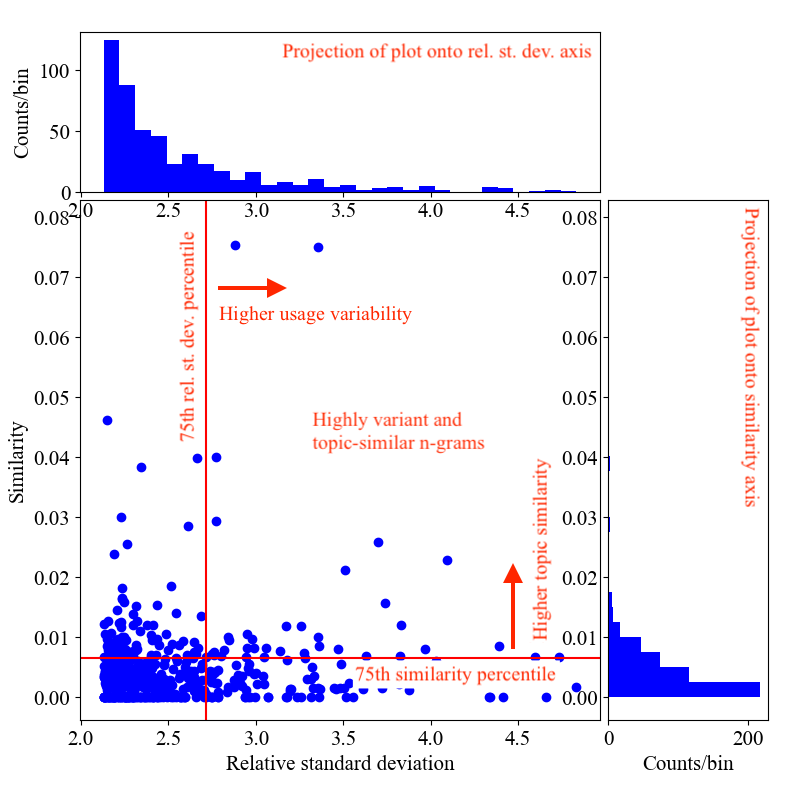}
\caption{Bivariate topic association measure for the “political events” topic. Each dot represents a unique $n$-gram. $n$-grams most likely to contribute to salience (horizontal) and used in context (vertical) with this topic are located in the upper-right quadrant defined by the red lines, marking the 75$^\mathrm{th}$ percentiles of the data distributed along each axis. Projections of each variable into one dimensional histograms are included for informational purposes.
}
\label{fig3}
\end{figure}
\FloatBarrier
\section{Quantifying dynamic topic salience}
The final step of this analysis requires quantifying the evolution of topic salience. Using topic $n$-gram sets identified via bivariate association and relative usage trends for each $n$-gram, we can determine the topic salience trends for each predefined topic. For a PMESII-ASCOPE topic, $T$, we defined a topic usage trend as the sum of the relative usage trends of all $n$-grams associated with $T$. We defined a topic salience trend as the average of the time derivatives of the relative usage trends of all $n$-grams associated with $T$. We define a salience matrix for time bin $t$ as the set topic salience values for all 36 PMESII-ASCOPE topics in a given time bin, or across a set of time bins.

Figure~4 presents monthly salience matrices during a four-month period from August 2017 to November 2017, covering both the general election and repeat general election. The months in which elections occurred (August and October 2017) feature high salience values especially for topics of a political nature (and especially the topic “political events”), but also for topics across the PMESII-ASCOPE space relative to non-election months, reflective of the multifaceted relevance of electoral politics to topics tracked by PMESII-ASCOPE.

Figure~5a revisits the topics ``political events'' and ``infrastructure capabilities'' discussed in Section 2 through the lens of temporal topic salience. In the same figure, we normalize salience values within each time bin; alternation of the two topics as most salient is visible in Figure~5b. As expected, we see high absolute and relative salience of ``political events'' surrounding the late 2017 election and we see salience of ``infrastructure capabilities'' in late 2016, suggesting the bivariate association routine described in Section 3 and our quantification of salience preserves trends in $n$-gram usage observed in Section 2. In addition to providing a measurement of how topic salience changes over time, particular distributions of salience values in the PMESII-ASCOPE topic space might signal the nature of recent, current, or upcoming events captured in a document corpus, at least in the scope of similar social, geographic, and historical contexts.

\FloatBarrier
\section{Discussion}
Figures 4 and 5 summarize the findings of this analysis relevant to an end user. Using these, we observe the relative salience of the whole topic space over the selected time frame (Figure~4) as well as the changing salience of particular topics across the entire chronology of the corpus (Figure~5). In Figure~4, we observe significant salience of the whole topic space, but notably of the ``political events'' topic, in the months surrounding the 2017 Kenyan elections; Figure~5's chronological representation corroborates this.

Using visualizations like that in Figure~5, an analyst or researcher working with a particular topic framework may observe (and discover previously unknown) patterns in the temporal salience of some topic or set thereof. Figure~4 represents the same data as Figure~5, but presents a holistic view of the semantic framework's topic salience distribution at specific points in time and may enable an analyst or researcher to discover notable relationships between the salience patterns of different topics.


\begin{figure}[hbt]
     \centering
     \begin{subfigure}[b]{0.49\textwidth}
         \centering
         \includegraphics[width=\textwidth]{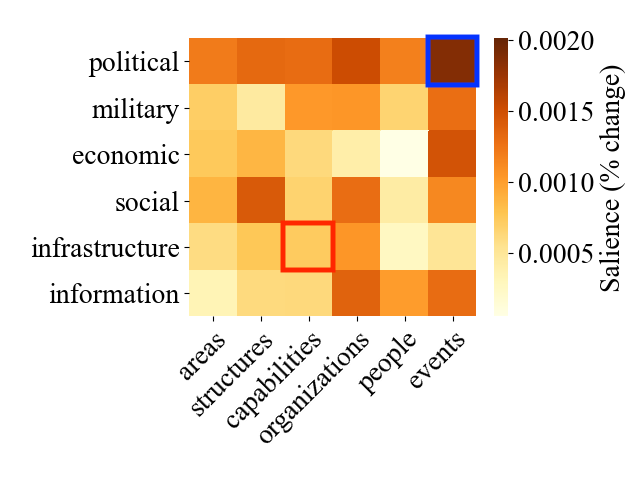}
         \vspace{-0.9cm}
         \caption{August 2017}
         \label{fig:fig4a}
     \end{subfigure}
     \hfill
     \begin{subfigure}[b]{0.49\textwidth}
         \centering
         \includegraphics[width=\textwidth]{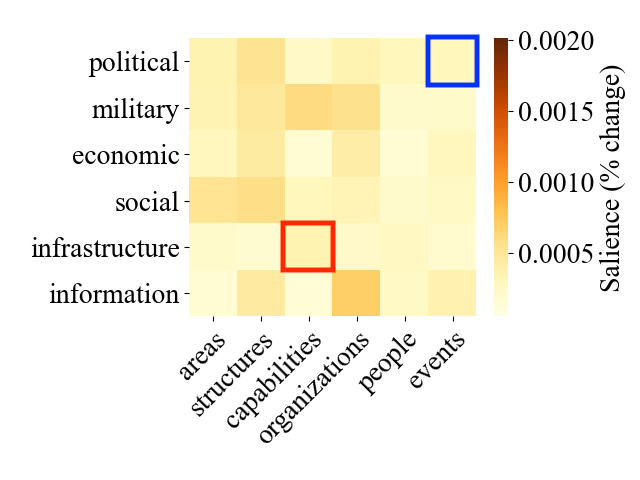}
         \vspace{-0.9cm}
         \caption{September 2017}
         \label{fig:fig4b}
     \end{subfigure}
     \hfill
     \begin{subfigure}[b]{0.49\textwidth}
         \centering
         \includegraphics[width=\textwidth]{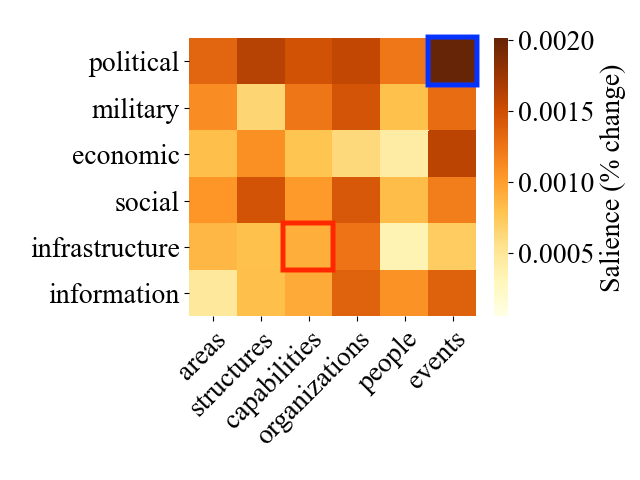}
         \vspace{-0.9cm}
         \caption{October 2017}
         \label{fig:fig4c}
     \end{subfigure}
     \begin{subfigure}[b]{0.49\textwidth}
         \centering
         \includegraphics[width=\textwidth]{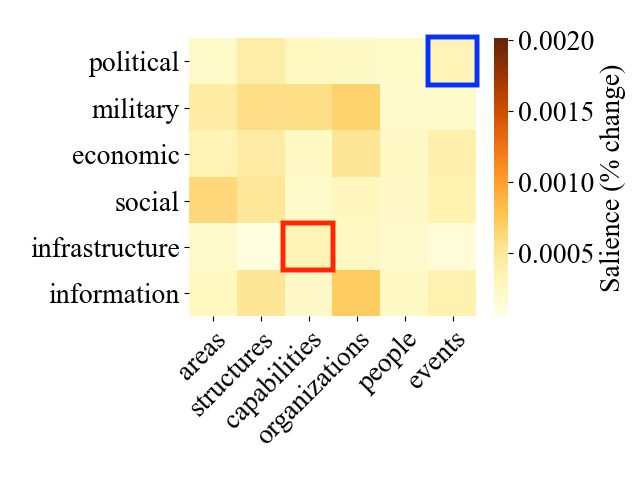}
         \vspace{-0.9cm}
         \caption{November 2017}
         \label{fig:fig4d}
     \end{subfigure}
        \caption{Salience matrices for four successive months of corpus documents showing snapshots of the relative salience of each of the semantic framework's topics. ``Political events'' (blue) and ``infrastructure capabilities'' (red) are highlighted; their trends are shown in Figure~5}
        \label{fig:fig4}
\end{figure}

\begin{figure}[htb]
     \centering
     \begin{subfigure}[b]{0.49\textwidth}
         \centering
         \includegraphics[width=\textwidth]{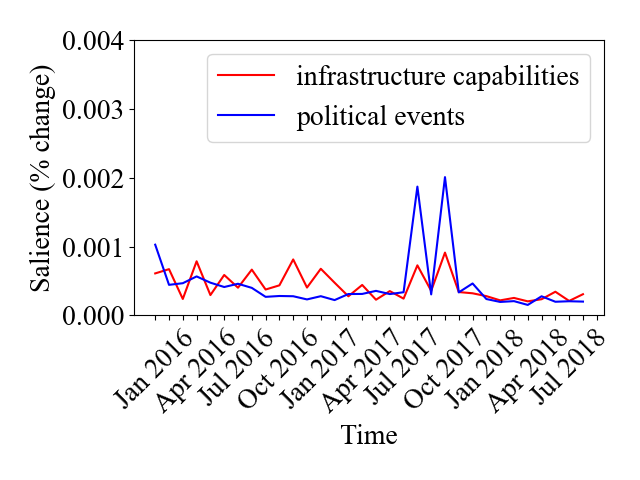}
         \vspace{-0.7cm}
         \caption{Absolute Topic Salience}
         \label{fig:fig5a}
     \end{subfigure}
     \hfill
     \begin{subfigure}[b]{0.49\textwidth}
         \centering
         \includegraphics[width=\textwidth]{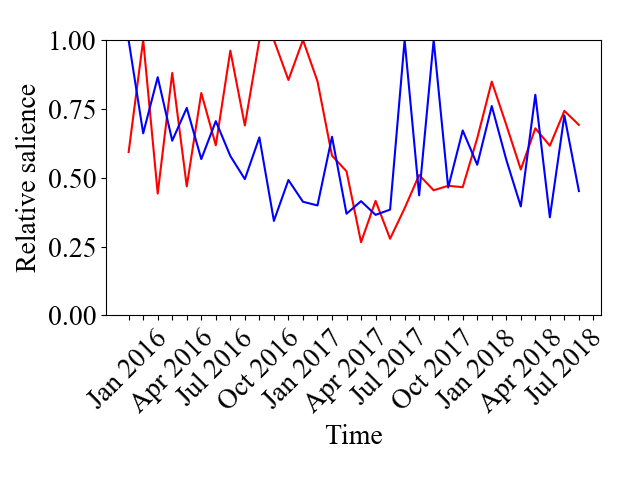}
         \vspace{-0.7cm}
         \caption{Relative Topic Salience}
         \label{fig:fig5b}
     \end{subfigure}
     \hfill
     \caption{Topic salience trends for topics ``political events'' and ``infrastructure capabilities''; (a) absolute salience and (b) relative salience (normalized with respect to the distribution of topic salience values at each time bin).}
     \label{fig:fig5}
\end{figure}

\FloatBarrier
\section{Conclusions}
\vspace{-0.15cm}
We present a novel and domain-adaptable approach to identifying dynamic salience of predefined topics of interest across a body of text. Potential applications include human, socio-cultural, and behavioral analytics and near real-time situation awareness, e.g., monitoring news or social media streams, to identify emerging salience of topics.

In addition to serving as a powerful analytical tool in such domains, our approach may serve to enhance existing traditional text mining. In Sections 1 and 2, we compare our approach with popular topic modeling techniques, including LDA. Since our approach is the epistemological reverse---using existing topics to identify patterns in content rather than building the topics themselves---it could be employed to validate or even extend the results of topic modeling, using generatively identified latent topics as an input topic framework whose salience patterns our technique would then seek to quantify.

Given an expectation that a chronologically organized corpus contains subject matter described by a predefined topic framework, our approach enables researchers and analysts to use a corpus as a basis for discovering how topics vary in salience over time and how they may correlate with one another. Our approach is agnostic to corpus and framework inputs whose realizations may span a large number of domains and applications.
 \vspace{-0.15cm}

\end{document}